# Survey on Reasoning Capabilities and Accessibility of Large Language Models Using Biology-related Questions

Michael Ackerman

School of Computing, The University of Georgia, Athens, GA 30602, United States

Email: ma23582@uga.edu
Phone: 404-307-0719

# Abstract

This research paper discusses the advances made in the past decade in biomedicine and Large Language Models. To understand how the advances have been made hand-in-hand with one another, the paper also discusses the integration of Natural Language Processing techniques and tools into biomedicine. Finally, the paper's goal is to expand on a survey conducted last year (2023) by introducing a new list of questions and prompts for the top two language models. Through this survey, this paper seeks to quantify the improvement made in the reasoning abilities in LLMs and to what extent those improvements are felt by the average user. Additionally, this paper seeks to extend research on retrieval of biological literature by prompting the LLM to answer open-ended questions in great depth.

# 1. Introduction

### 1.1 Purpose

The goal of this survey is to quantify the improvement made in reasoning capability between the two leading Large Language Models (GPT3.5 and GPT4). Additionally, this survey will explore the extent to which the use of prompt engineering affects each model in order to determine whether the improvement in reasoning capacity between the two models will be felt by the average user. Effectively testing the improvement in reasoning capacity and accessibility that was made between the GPT3.5 generation and the GPT4 generation of Large Language Models. The goal is to test the capabilities of these generations of LLMs at their peak; toward this goal GPT3.5 and GPT4 make for the most suitable comparison. This survey is meant to serve as an extension to a survey conducted in 2023 (Gong et al., 2023), where GPT3.5 and GPT4 were found to be the leading models in reasoning-based questions. This survey will aim to: expand the research conducted by the 2023 group by introducing open-ended questions as well as a grading rubric, and remove extraneous variables by only two prompts and models to evaluate the improvement. Additionally, while the 2023 survey aimed to find the model most suitable for reasoning, this survey is conducted under the pretense that GPT4 is the leading Large Language Model. The question I am attempting to answer through this survey is "How much have reasoning-capabilities improved between the last two generations of LLMs and will that improvement be accessible to the average user"?

### 1.2 Reasoning in LLMs

Reasoning is a rising topic in Large Language Models requires the breaking down of a complicated, reasoning-based problem by the LLM into smaller subproblems and putting the information back together in order to solve the overall problem. This topic is particularly significant to the average user, who is likely to ask reasoning-based questions and expect a straight-forward answer in response. A common issue with earlier generations of LLMs was their inability to give straight-forward answers, often avoiding the question and simply presenting the user with loosely-explained bits of information regarding the topic. In order to address this issue and quantify the improvement made in the quality of responses to reasoning-based questions, I will employ two methods. The first method I will use comes down to question design; as mentioned previously, I will be using open-ended questions. One reason for this is to expand on the robustness of the 2023 survey on large language models, however, another reason is for testing the LLM's ability to answer a complex, open-ended, question without deflecting through the means mentioned above. Additionally, the scoring rubric heavily penalizes a lack of relevant information, whereas, an excess of irrelevant information is not as heavily punished (potentially all points lost for not enough relevant information and only 1 of 3 points potentially lost from too much irrelevant information). This rubric is specifically designed to reward the model for giving more relevant information, as well as putting it together in a more interpretable way. Therefore, the rubric design is another factor specifically designed to test the model's reasoning abilities in an open-ended scenario.

### 1.3 Prompt-Engineering

Prompt-engineering is the process of creating a prompt for a language model that is specifically designed to optimize its use and get the best result from that language model. Prompt engineering leads to far more reliable and relevant results than simply talking to the language model the same way one would normally speak. The improved performance of LLMs when using prompt engineering makes it a valuable skill for anyone using language models. However, the average user of LLMs (such as ChatGPT aka. GPT3.5) does not know how to properly engineer a prompt. Therefore, to simulate the experience of an average user (specifically the improvement in the average user's experience) in this survey, I will need to quantify the difference in the quality of LLMs' responses using an engineered prompt against their responses using a non-engineered prompt. A smaller difference in GPT4's scores when using the engineered prompt as compared to the non-engineered prompt than GPT3.5's scores when using the engineered as compared to the non-engineered prompt would mean that GPT4 has become more accessible than GPT3.5 was. This is because the difference between GPT4's responses would be less influenced by the prompt, effectively meaning that the Language Model gives accurate, robust, and reliable answers to prompts that could be generated by the average user (such as natural speech perhaps). This could potentially be achieved by making more assumptions about the user's query. Therefore, ensuring they get the information they were looking for (without being a prompt-engineering expert), at the risk of getting more information than they asked for.

## 2. Related Works

### 2.1 Advancements in Biomedicine

In the past decade, biomedicine has advanced at a high rate, with many existing technologies being incorporated into the field and many new technologies being invented specifically for the field of biomedicine. A study conducted in 2016 (Fernandez-Leiro et al., 2016) explored the future of Cryo-electron microscopy and the challenges to mapping the three-dimensional structure of macromolecules. CryoEm (cryo-electron microscopy) allows biological samples to be kept at cryogenic temperatures, which "enables their preservation in a high vacuum and provides them with some protection against the effects of radiation damage"(Fernandez-Leiro et al., 2016). In their analysis of the CryoEm technology, the group detailed many areas in which improvement could lead to drastically better results and a wider variety of uses for the technology. Of those areas one stood out the most:

> "The ability to obtain structures from small subsets of the data through image processing, possibly in combination with sample preparation on the nanolitre scale, will also expand the applicability of cryo-EM to complexes with low abundance in the cell. Using time-resolved methods, even transient complexes and intermediate conformational states can be studied, providing unprecedented insight into the function of large macromolecular machines. For example, the structural characterization at the molecular (or even the atomic) level of extremely large and flexible machines such as the nuclear pore complex[122], as well as the characterization of complexes involved in the organization of chromatin[123], is now on the horizon."(Fernandez-Leiro et al., 2016)

This study showed that the integration of technology, specifically imaging and dataset manipulation technology, into the biomedical field was on the rise in 2016. Thus defining the need for the integration of an existing technology into, or the creation of a new technology for, the biomedical field. This new technology would need to be able to process image data, recognize patterns, and modify image-datasets in order to accomplish these tasks with very minimal real data. In this paper, the research group had defined a need for "few-shot" technology in the pursuit of three-dimensional mapping of macromolecules. "Few-shot" technology refers to a kind of machine learning framework that trains language models to make accurate predictions using minimal data.

Three years after the CryoEm paper, another paper (Yin et al., 2019) discussing the future of a new, potentially revolutionary, technology in the biomedical field was published. The paper discussed CRISPR-Cas:

> "an RNA-guided, targeted genome-editing platform with great potential in both basic research and clinical applications … particularly … in cancer research and oncology drug development" (Yin et al., 2019)."

CRISPR systems are a far advancement from the first "technologies using zinc-finger nucleases (ZFNs)", which were "efficient" but required "a fairly complicated process of protein engineering … to target specific DNA sequences" (Yin et al., 2019). The second generation of genome-editing platforms used "transcription activator-like effector nucleases (TALENs) for efficient genome editing" (Yin et al., 2019). Both of these systems "can introduce DSBs" (double stranded breaks), "which are repaired by NHEJ or HDR", however, "the design and assembly process of TALENs is faster than those of ZFNs, and the potency and specificity of TALENs are potentially higher too" (Yin et al., 2019). The third generation of genome-editing platforms are CRISPR systems, which are "predicated on RNA-guided nucleases"; additionally, "class 2 CRISPR systems consist of a Cas endonuclease and at least one target-specific CRISPR RNA (crRNA)" (Yin et al., 2019). CRISPR systems are even more specific than TALENs as they use "the PAM sequence", which "enables distinction between self DNA versus foreign DNA" (Yin et al., 2019). Additionally, CRISPR systems provide a wider variety of uses through a multitude of systems such as "*Streptococcus pyogenes* Cas9 (SpCas9)" as well as "Cas9 analogues, such as *Streptococcus thermophilus* Cas9 (StCas9) or *Neisseria meningitidis* Cas9 (NmCas9)" and "*Staphylococcus aureus* Cas9" (Yin et al., 2019). Moreover, CRISPR systems are also able "to regulate the expression of a target gene" (Yin et al., 2019) through the use of CRISPR activators and inhibitors (CRISPRa and CRISPRi). Lastly, CRISPR systems have shown to be a groundbreaking innovation in the diagnostics and treatment of cancer due to their RNA/DNA mismatch detection enzyme Cas13a being "sensitive enough to detect cancer mutations present in a concentration as low as 0.1% of total DNA" (Yin et al., 2019).

This paper, titled "CRISPR–Cas: a tool for cancer research and therapeutics", defined an area of progress for CRISPR systems in reducing "off-target genome editing" (Yin et al., 2019). The research group states that "the results of genome-wide off-target analysis suggest that the experimental conditions can be optimized to obtain a low level of off-target activity with substantial on-target editing" (Yin et al., 2019). Optimizing the experimental conditions is a task that could be pursued using Large Language Models by running simulations using available data and modifying the data using dataset manipulation techniques (similar to the "few-shot" approach discussed previously).

Following the paper on CRISPR-cas, a paper titled "Multilevel omics for the discovery of biomarkers and therapeutic targets for stroke" (Montaner et al., 2020) was published. This paper discussed "how multi-omics techniques are contributing to the discovery of biomarkers for diagnosis and prognosis in ischaemic stroke" as well as the use of these techniques "to identify molecular targets for therapeutic interventions" (Montaner et al., 2020). Differentiating between Ischaemic stroke and stroke caused by intracerebral hemorrhage rapidly is crucial in treating stroke. In the pursuit of making this distinction, identifying certain biomarkers (such as "glial fibrillary acidic protein (GFAP)… retinol-binding protein 4 (RBP4)... matrix metalloproteinase 9 (MMP9)", etc.) and their levels in the bloodstream in a "pre-hospital setting" could lead to the ruling out of certain diagnoses and faster treatment (Montaner et al., 2020). "Multi-Omics" refers to the "use of high-throughput techniques based on large screening processes to avoid selection bias and … [generate] extensive lists of molecules for evaluation as biomarkers" (Montaner et al., 2020). Of the multiple "omics-based approaches" ("proteomics … genomics, transcriptomics and metabolomics"), the most commonly used "tool for discovering new biomarkers" is proteomics (Montaner et al., 2020). The proteomic approach has been in use since 2009, when "the first study of brain proteomics in patients with stroke" used "laser microdissection … to investigate MMP expression profiles in the neurons and vasculature of the ischaemic brain" (Montaner et al., 2020). This approach was so successful and developed so rapidly that "by 2013, 51 proteins had been identified as differentially expressed in the infarcted brain after ischaemic stroke" (Montaner et al., 2020). Those proteins included GFAP, the protein previously discussed as one of the leading biomarkers in distinguishing Ischaemic stroke and stroke caused by ICH, as well as " Other lesser-known proteins, such as *N*-ethylmaleimide-sensitive factor (NSF) ATPase, which is critical for membrane trafficking in neurons" (Montaner 2020). However the sources of biomarkers for strokes (as well as the different types/causes of stroke) don't end there as "In addition to studies of brain proteomics, protein expression patterns in accessible bodily fluids have also been studied to better understand the pathophysiology of ischaemic stroke" (Montaner et al., 2020). Given the large amount of information that could be used to diagnose a stroke and distinguish the type of stroke, as well as the significance of identifying a stroke and its cause rapidly, the paper presents a potential area of improvement in this field:

> "The progress of all omics technologies has necessitated development of tools to facilitate analysis and interpretation of the multidimensional data being generated. Many statistical methods have been developed for independent analysis of large-scale, high-quality data from each level of omics, but such individual interpretations overlook the crosstalk between different molecular entities and could miss biologically relevant information. Consequently, integrated analysis of data obtained with different omics approaches — here referred to as integromics — is becoming crucial for a deep understanding of pathological processes in a biologically meaningful context." (Montaner et al., 2020)

In the pursuit of identifying the combinations of biomarkers from different parts of the body that could signal an oncoming, in-progress, or recent stroke the research group defines a need for a technology with particular capabilities. This technology would have to be able to conduct statistical analyses of "large-scale, high-quality data from each level of omics" (Montaner et al., 2020), which is a task highly suited for Large Language Models (or technology that is predicated on large language models, specifically transformers). As biomedicine has advanced in the past decade, the need for more powerful data-processing and dataset-augmenting technology has arisen in order to train algorithms/models to recognize certain signs of diseases/disorders and allow for faster, more accurate treatment. Increasingly, the capacity to fill the need for such a technology has become more accessible with the rapid improvement of Large Language models, which are predicated on the concepts of dataset manipulation, data analysis and statistical approaches to data concatenation.

## 2.2 Advancements in Language Models

Large Language Models have been on the rise for nearly a decade and have been improving at an exponential rate since their inception. In this section, I will specifically cover their improvement in natural-language processing for a variety of purposes.

Last year (in 2023) a paper, titled "AugGPT: Leveraging ChatGPT for Text Data Augmentation", was published discussing "Text data augmentation[;] … an effective strategy for overcoming the challenge of limited sample sizes in many natural language processing (NLP) tasks" (Dai et al., 2023). This paper discussed data augmentation, "the artificial generation of new text through transformations, is widely used to improve model training in text classification" and some techniques for augmentation "at different granularity levels: characters, words, sentences, and documents." (Dai et al., 2023). Some techniques for augmentation on the character level include "randomly inserting, exchanging, replacing, or deleting of characters in the text" (Dai et al., 2023). Additionally, there are: "optical character recognition (OCR)", which "generates new text by simulating the errors that occur when using OCR tools to recognize text from pictures", spelling augmentation, which "deliberately misspells some frequently misspelled words" and Keyboard augmentation, which "simulates random typo errors by replacing a selected key with another key close to it" (Dai et al, 2023). On the word level, the techniques include: "random swap augmentations", "random deletion augmentations", "synonym augmentation", "word embedding augmentation" (and its improved version, "the counter-fitting embedding augmentation"), and "contextual augmentation" (Dai et al., 2023). For the sentence and document levels of granularity, text data augmentation techniques include "back translation", which "uses translation models for data augmentation", as well as "[paraphrasing] the entire document to preserve document-level consistency" (Dai et al., 2023). These techniques are specifically useful for "few-shot learning scenarios", which are scenarios where data is either limited, low quality, or both (methods include "prompt-tuning" and "meta-learning"). Specifically, data augmentation is most useful for "very large language models", which are "Pre-trained language models (PLMs) based on the transformer architecture (such as the BERT and GPT)" (Dai et al., 2023) that use Natural Language Processing techniques. For this paper, the research group used three different datasets (amazon reviews for 24 different products, symptoms dataset from kaggle, and the PubMed20K dataset to create a framework based on a "base dataset" (all labeled samples) and a "novel dataset" (very few labeled samples, reminiscent of few-shot scenario). After BERT (Bidirectional Encoder Representations from Transformers) is fine-tuned on the base dataset, the research group uses ChatGPT to augment the data and fine-tunes BERT again on the augmented dataset (BERT is google's language representation model). "ChatGPT is regarded as an unsupervised distribution estimation" in pre-training and uses "Reinforcement Learning from Human Feedback (RLHF) to fine-tune the pre-trained language model" (Dai et al., 2023). Afterwards, ChatGPT goes through: "Supervised Fine-tuning (SFT) … Reward Modeling (RM) … " and "Reinforcement Learning (RL):" (Dai et al., 2023).

Due to the pretraining "on large-scale corpora", the introduction of " a large number of manual annotation samples" from the fine tuning stage, and the high quality data generated through the reinforcement learning, ChatGPT was determined to be the optimal tool for data augmentation in this study. The results of this study found that AugGPT had the highest labeling accuracy in all three datasets (amazon, symptoms, PubMed20K). Data augmentation increased accuracy in the PubMed20K dataset by about 4% (79.2/79.8 - 83.5), and accuracy in the symptoms dataset was increased by nearly 30% (63.6/60.6 - 88.9/89.9). Thus demonstrating that for the purposes of data augmentation for few-shot scenarios (such as the scenarios described above in the CryoEm paper (Fernandez-Leiro et al., 2016)), Large Language Models (especially ChatGPT) are a suitable tool to fill the need for image-dataset augmenting technologies.

More recently (2024), a paper (Chen et al., 2024) titled "Improving Code Generation by Training with Natural Language Feedback" was published. This paper explores the improvement in Language Models' abilities to synthesize programs; "program synthesis … [is] the automatic generation of computer programs from an input specification" (Chen et al., 2024). This paper is meant to solve the issue of LLMs "[struggling] to consistently generate correct code, even with large-scale pre-training", which the research group attributes to the methods used for filtering data for the "code pretraining datasets" (Chen et al., 2024). To address this issue, the group developed "Imitation learning from Language Feedback (ILF)", which is a "user-friendly … ", and "sample-efficient" "algorithm for learning from natural language feedback at training time" (Chen et al., 2024). For their experiment, the group "[trained] and [evaluated] [their] models on the Mostly Basic Python Problems (MBPP) dataset" (Chen et al., 2024). The research group also "[used] the human feedback annotations to create few-shot feedback prompts", effectively evaluating their model's ability to "Incorporate feedback" and synthesize programs in few-shot scenarios (Chen et al., 2024). The results this group found were that the only method with comparable pass rates to ILF with "$\pi_{Refine}$ refinements" (36% pass @1 rate and 68% pass @10 rate) was Gold Standards with "human refinements" (33% pass @1 rate and 68% pass @10 rate). Where $\pi_{Refine}$ is an algorithm "that is already fine-tuned or can generate refinements via few-shot prompting" (Chen et al., 2024). According to the research group:

> "ILF outperforms both finetuning on MBPP gold programs and human-written refinements on the pass@1 metric, yielding 14% absolute (64% relative) and 3% absolute (9% relative) increases in pass@1 rates, respectively"
> (Chen et al., 2024).

This paper demonstrates that ILF is an exciting new innovation in Natural Language Processing and that "ILF can significantly improve the quality of a code generation model" (Chen et al., 2024). The techniques of combining real-time human feedback (which the model was shown to have used to some extent) as well as dataset augmentation and few-shot feedback prompts to generate few shot scenarios could be useful in all areas of development for Large Language Models. Additionally, as the research group stated, ILF "is not model-specific … and can be conducted in multiple rounds to continuously improve the model" (Chen et al., 2024). Specifically, this group's design of their training datasets in combination with the imitation learning is reminiscent of the technology that was shown to be needed in the multi-omics paper (Montaner et al., 2020). The dataset filtering techniques would allow for better filtering of relevant and irrelevant biomarkers (this could also be helpful for advancing CryoEm (Fernandez-Leiro et al., 2016), as 3-dimensional imaging tends to result in a lot of noise), and the imitation learning could help a model learn what combinations of these biomarkers are signals of a stroke.

## 2.3 Integration of LLMs in Biomedicine

The effort to integrate language models (as well as other information processing techniques) into the biomedical field has been prevalent for over a decade. Specifically, the field of bioinformatics has seen much progress due to the integration of technologies that use natural language processing as well as image processing and are able to detect patterns that are not visible (or easily detectable) to humans, very reliably. Additionally, certain Natural Language Processing techniques can be used in biomedicine to analyze data that is very limited and of low quality by augmenting datasets to mimic large quantities of high-quality data. All of these advancements in Natural Language Processing have drastically increased development in the field of bioinformatics.

When Language Models were first being introduced into the field of bioinformatics, a paper titled "Survey of Natural Language Processing Techniques in Bioinformatics" (Zhiqiang et al., 2015), which discussed the integration of Natural Language Processing techniques in bioinformatics research. Due to the nature of bioinformatics, "Informatics methods, such as text mining and natural language processing, are always involved in bioinformatics research" (Zhiqiang et al., 2015). According to the research group, "Bioinformatics is an interdiscipline that emerged with the progress and accomplishment of the Human Genome Project" and "Data storage, retrieval, and analysis are the key processes in bioinformatics" (Zhiqiang et al., 2015). There are many uses for text processing techniques in Bioinformatics; including, "text mining technology" for "retrieving biological literature" and "establishing biological information databases" (Zhiqiang et al., 2015). Additionally, "with the exponential growth of biological literature, a program that can recognize protein-protein interactions automatically from PubMed abstracts" (Zhiqiang et al., 2015). Another use of Language Processing Techniques is "extracting the relationship between gene functions and diseases", which "involves searching for gene names and disease names simultaneously in the literature and then determining whether a particular gene is related to a certain disease" (Zhiqiang et al., 2015). Lastly, Natural Text Processing techniques can be used to "obtain answers to many … bioscience and bioinformatics problems in various databases, such as PubMed." (Zhiqiang et al., 2015). Natural Language Processing techniques can be used for predicting protein structure and function, predicting RNA mismatches through comparative and non-comparative methods, and many more potential purposes as their capabilities improve. "The development of bioinformatics relies on information science" and "natural language processing researchers should provide a more extensive application space" (Zhiqiang et al., 2015).

More recently, in 2022, a paper, titled "ProtGPT2 is a deep unsupervised language model for protein design" (Ferruz et al., 2022) was published. This paper discussed ProtGPT2, "an autoregressive Transformer model with 738 million parameters capable of generating de novo protein sequences in a high-throughput fashion." (Ferruz et al., 2022). The research group claims that "The major advances in the NLP field can be partially attributed to the scale-up of unsupervised language models … ", which "do not require annotated data" (Ferruz et al., 2022). In their experiment, the research group used a transformer to create a generative language model, which they trained using an "autoregressive strategy",  to generate protein sequences. The group used the "UniRef50 (UR50) (version 2021_04)" dataset in the training and "randomly excluded 10% of the dataset sequences" (Ferruz, 2022) from the training dataset in order to later use those sequences for evaluation. In their experiment, the group first ran IUPred3 on the protein level to analyze how prone ProtGPT2-generated sequences were to being disordered. They found that ProtGPT2-generated sequences had a similar ratio of ordered/disordered sequences to the set of natural sequences (87.59% and 88.40% ordered sequences, respectively) (Ferruz et al., 2022). Additionally, on the amino-acid level, the group found that ProtGPT2 sequences and the set of natural sequences had a "similar distribution of ordered/disordered regions" (79.71% and 82.59% ordered sequences, respectively) (Ferruz et al., 2022). The group's research found that "ProtGPT2 generates sequences that resemble globular domains whose secondary structure contents are comparable to those found in the natural space." (Ferruz et al., 2022). This research demonstrates that the generative capabilities of ProtGPT2 could be used to substantially increase the rate of development in the fields of 3-dimensional protein mapping, protein-to-protein interactions, and the discovery of new proteins for use in the biomedical field. The group concluded that "Since protein design has an enormous potential to solve problems in fields ranging from biomedical to environmental sciences … ", "ProtGPT2 is a timely advance towards efficient high-throughput protein engineering and design." (Ferruz et al., 2022), which could have a large variety of applications in biomedicine.

Lastly, just last year (2023), a paper, titled "Evaluating the Potential of Leading Large Language Models in Reasoning Biology Questions", was published. For this paper, the group evaluated the reasoning capabilities of 5 Large Language Models using a list of 108 multiple-choice questions (all related to biology) and evaluating the total scores, confidence levels, and standard deviations of each model's responses. The results this group received will be discussed in the methodology section as the survey conducted in their 2023 paper will serve as the basis for the survey I will conduct in this paper. I will extend their evaluation of the top two performing models (GPT3.5 and GPT4) using open-ended questions, two prompts (one that is engineered and one basic prompt), and a standardized scoring rubric. My goal is to extend the previous paper's findings on the improvements made in reasoning capabilities between GPT3.5 and GPT4 as well as the accessibility of each LLM.

# 3. Methodology

## 3.1 Model Evaluation

In this survey, the two models I chose to use for evaluation were GPT3.5 and GPT4. There are multiple reasons I chose to compare the older model (GPT3.5) with its own, newer (pay-walled) version (GPT4). The first reason for this comparison is that it allows me to directly measure the overall improvement made between the two models in regard to reasoning in general. Additionally, ease of use becomes more quantifiable as the techniques for engineering prompts are presumed to be standardized across the two models. The expectedly similar styles of input, as well as output, allow me to evaluate the score more closely by comparing the amount of points in each category. This would be more difficult to evaluate accurately with models that have drastically different styles of responses. Therefore, assuming the most similar of all models are GPT3.5 and GPT4 makes their evaluation more accurate. Lastly, evaluating these two models alone allows me to accurately measure the improvement made in general reasoning, as well as ease of use between different generations of language models.

Furthermore, a survey conducted in 2023 (Gong et al., 2023) resulted in the following data:

> GPT4 was the highest performing model (of GPT4, GPT3.5, PaLM2, Claude2, and SenseChat/SenseNova). Receiving the highest average score (90/108) along with the lowest standard deviation (0.30) and highest average correlation (0.87); showing that GPT4 is more accurate, consistent, and confident with its answers. Surprisingly, the model that performed second best in most categories was GPT3.5, receiving the second highest average score (80 for GPT3.5 compared to 79 for Claude2 in third). Additionally, GPT3.5 had the second lowest standard deviation by yet another small margin (0.40 for GPT3.5 compared to 0.41 for Claude2). Meaning that, on average GPT3.5 and GPT4 outperformed all other models tested in the survey.

In order to test the capability of large language models to answer reasoning based questions well without special prompts, it is most efficient to single out the two highest performing LLM's (GPT3.5 and GPT4) in the reasoning category. Setting the standard for accuracy as high as possible allows me to further separate the model's general ability to answer the question and the model's ability to answer the question given the correct prompt.

GPT3.5 and GPT4 are different generations of the same large language model, making their accuracy, as well as their usability much easier to evaluate. Additionally, they were shown to be the two most accurate models in the category of reasoning-based questions (specifically with regard to biology), which is the topic of this paper. Therefore, in order to most accurately evaluate LLMs at their current peak in reasoning, these are the two most fitting models. Lastly, evaluating different generations of the same model also allows me to evaluate the improvement made between generations of LLMs, making this survey more significant.

## 3.2 Question Design

For the survey, I designed a list of 50 open-ended questions to test the reasoning capabilities of a large language model. Of the 50 questions, I wrote 25 of them by hand and designed them to be complex and thought provoking. Five of the hand-written questions focused on ethics, sometimes presenting the model with a hypothetical scenario and asking it for advice or a solution. The other 20 hand-written questions asked the model to describe complex biological processes, or come up with creative solutions to hypothetical questions. In this survey, I also used 25 questions that were generated by GPT4 designed to test general knowledge in biology-related topics. The List of questions can be found here: [Questions](#)

## 3.3 Prompt Design

There are two prompts used in this survey. They are as follows:

Prompt 1: "Answer the following questions as if asked individually by a user, without any further context"

Prompt 2: "Please answer the following practice questions as if you were a resident in training for a biological exam. Answer each question with a full paragraph, giving a detailed response. Ensure that your response to each question provides a satisfactory answer that would receive a good grade on the exam."

The first Prompt is designed to test the model's ability to respond correctly, with depth, in an interpretable manner without the use of prompt engineering. In this survey, I will use the simple prompt to simulate an average user. On the other hand, the second prompt is designed using prompt engineering techniques. Such techniques include: providing the model with a perspective from which to answer the questions (a resident…), setting a minimum length for each answer (a paragraph), and emphasizing that each answer would have to receive a good grade. Such a prompt covers every section of the scoring rubric for this survey (accuracy, depth of content, and interpretability). Therefore, this prompt is used as a control group to test how well the model would answer the questions when used in a more sophisticated manner.

## 3.4 Test Procedures/Scoring Analysis

For this survey, each of the two models were asked a series of 50 questions (provided in 3.2), after being given the two different prompts. Each prompt was followed by the full list of questions, in the same message. With the case of GPT4, it would stop generating after answering 3-4 questions and provide a pop-up to allow the user to "continue generating" from the point at which it last stopped. This mechanic was used to solicit the full list of responses from GPT4. In the case of GPT3.5, due to the character limit, the option to "continue generating" was not provided as a pop-up, and the model would occasionally end its responses with the phrase "(Continued below due to character limit)". In this case, the model was manually prompted to "continue generating" via the chat box until it finished answering all of the questions. Both models were good at remembering the questions between different character limit checkpoints. However, GPT3.5 started to deviate from the list of questions around question 47 and needed to be prompted again, with the correct questions.

Once the responses were recorded, they were stored in groups and graded by hand using the following rubric: (with brief justification provided on the answer docs)

"Scores will be given for each answer in a range from 0-3 where each point given depends on the previous point. The first point is given for correctness of the answer, if no point is given here no points can be granted for the answer. The second point is for depth of content, if no point is given here, the answer earns one point. The third point is given for ease of interpretability, if no point is given here, the answer receives 2 points, if this point is granted, the answer receives the maximum of 3 points. This is done because if the answer is not accurate, the depth of its content is not relevant, and if the answer does not have much depth/complexity, then interpretability is assumed to not be an issue."

For simplicity, I will refer to these groups as GPT3.5-1, GPT3.5-2, GPT4-1, and GPT4-2; where each group name consists of the name of the model being tested followed by a dash and the number of the prompt being used in each particular test.

In order to avoid extraneous variables, neither model got to see the rubric on which it would be graded as that was assumed to have a similar effect as engineering a specialized prompt.

## 3.5 Result Analysis and Expectations

As mentioned above, I divided the responses into groups of 4. After the scoring is complete, the scores will be analyzed in three different groupings. Firstly, I will compare all GPT4 responses against all GPT3.5 responses in order to understand the overall improvement made between the two models (using equal weight for prompts that are engineered and prompts that are not). To quantify the overall improvement, I will compare the total points of GPT3.5 and GPT4 responses against each other. The second grouping I will explore in this survey is not meant to compare the models but rather the extent to which each prompt affects the individual models. Specifically, I will compare the total points of GPT3.5-1 and GPT3.5-2 in one section; I will look at the total points of GPT3.5-1 and GPT3.5-2 separately. The reason for this is so that I can quantify the level of significance prompt engineering holds in GPT3.5 and compare that to the significance of prompt engineering in GPT4. Lastly, I will compare the points either model received in each individual category. For this last comparison, I will only look at the second prompt, additionally, the scoring for each category will be relative to the current pool of available points. Ex: If GPT3.5-1 receives a total score of 20 in the accuracy category, the score it will receive in this comparison for the depth category will be a percentage of 20. I chose to compare the data in this way so that I can pinpoint the category in which the most improvement has been made between the two models.

On the basis that GPT3.5 is the predecessor of GPT4, I can expect certain outcomes in this survey. My expectations for the first comparison are:

1. GPT3.5: Out of the total amount of points available to each model (300) I expect GPT3.5 to be within a range of 66-88.
2. GPT4: I expect GPT4's total points to be within a range of 209-244.

My expectations for the second comparison are (150 total points for each model-prompt combination):

1. GPT3.5-2 will outperform GPT3.5-1 by a significant margin (range of 20-30 points).
2. GPT4-2 will also outperform GPT4-1 by a similar range (20-30 points). However, relatively speaking this range is far less significant given the total points I expect GPT4 to earn. Most of the discrepancy likely arises from the interpretability section with some points lost in the depth section as well.

For the third comparison, I will only define my expectations for each model in each category. However, once the survey is complete, I will also analyze the data for each model-prompt combination in the results section. The being said, my expectations for the third comparison are (100 points in each category for each model):

1. GPT3.5: I expect GPT3.5 to be within a relatively high range of 60-80 points in the accuracy category. However, I expect only 10% of accurate answers (6-8) to receive a point for depth. Additionally, I expect no points for this model in the interpretability category.
2. GPT4: I expect GPT4's points in the accuracy section to be within a range of 95-100. Additionally, I expect 80-90% of the accurate answers (76-90) to receive a point for depth. Interpretability is where I expect the most loss with only 50-60% of all deep answers (38-54) receiving a point for interpretability.

# 4. Results

In order to properly express the scores and make this survey repeatable I will provide each model's responses (for each of the two prompts) with scoring and justification included: GPT3.5-1, GPT3.5-2, GPT4-1, GPT4-2 as well as the document containing all tables used in this paper: Grading

## 4.1 Comparison Between GPT3.5 and GPT4 and Intra-model Comparison Between Prompts

For the first, inter-model, comparison I will look at the totals of the following two tables:

| GPT3.5   | Section 1(Out of 50) | Section 2(Out of 50) | Section 3(Out of 50) | Totals  |
|----------|----------------------|----------------------|----------------------|---------|
| Prompt 1 | 29                   | 4                    | 0                    | 33/150  |
| Prompt 2 | 45                   | 41                   | 21                   | 107/150 |
| Totals   | 74/100               | 45/100               | 21/100               | 140/300 |

GPT3.5 Score Breakdown

| GPT4     | Section 1(Out of 50) | Section 2(Out of 50) | Section 3(Out of 50) | Totals |
|----------|----------------------|----------------------|----------------------|--------|
| Prompt 1 | 34                   | 6                    | 2                    | 42/150 |
| Prompt 2 | 49 | 44 | 27 | 120/150 |
| Totals | 83/100 | 50/100 | 29/100 | 162/300 |

## GPT4 Score Breakdown

GPT3.5 had scored a total of 140 points, drastically outperforming my expected range for this model (66-88 total points) scoring 52-74 more points than expected. On the other hand, GPT4 only scored a total of 162 points, underperforming my expected range (209-244) by a significant margin (47-80 points). From my observation of the data, it is clear that this discrepancy between my expectations and the reality of the scores largely resulted from the fact that I overestimated the capabilities of GPT4 when using a non-engineered prompt (GPT4-1 scoring a total of 42/150 points). Additionally, I drastically underestimated the capabilities of GPT3.5 when using an engineered prompt (GPT3.5-2 scoring 107/150 points). This finding demonstrates that the improvement in reasoning capabilities between GPT3.5 and GPT4 is not very drastic, with GPT3.5-1 scoring only 9 points less than GPT4-1 (33 and 42, respectively) and GPT3.5-2 scoring only 13 points lower than GPT4-2 (107 and 120, respectively). The further implications of this data will be discussed in the discussion section, however, this data does imply that GPT4 is not significantly better than GPT3.5 at reasoning.

Similarly to the first comparison, I will be using the above tables for the second comparison. My observations for each of the models were as following:

> GPT3.5: I expected GPT3.5-2 to only outperform GPT3.5-1 by 20-30 points, which would have been a large margin given my expectations for the overall performance of GPT3.5. In my data, I observed a 74 point discrepancy between GPT3.5-1 and GPT3.5-2 (33 and 107, respectively).

> GPT4: I expected GPT4-2 to outperform GPT4-1 by a similar margin to the the one found in the data for GPT3.5 (20-30 points). In my data, I observed a 78 point Discrepancy between the GPT4-1 and GPT4-2 (42 and 120 points, respectively).

Given that my data suggests that difference in prompt-types accounts for over double the discrepancy I expected to see in the two sections, this data clearly demonstrates the persistent significance of prompt-engineering in the use of Large Language Models. Seeing as the model accounted for far less of a point discrepancy than expected and the prompts accounted for a far greater point discrepancy than expected, it could be reasoned that the improvement made in reasoning between GPT3.5 and GPT4 is mostly felt when using engineered prompts. The larger discrepancy between the two models when using the engineered prompt as compared to the non-engineered prompt (13 point discrepancy using an engineered prompt; 9 point discrepancy using a non-engineered prompt) would further support this notion.

## 4.2 Comparison in Each Scoring Category

For the third comparison, I will use three different tables to look at each point category individually:

| Accuracy | GPT3.5 | GPT4 |
|---|---|---|
| Prompt 1(Out of 50) | 29/50 = 58% | 34/50 = 68% |
| Prompt 2(Out of 50) | 45/50 = 90% | 49/50 = 98% |
| Totals | 74/100 = 74% | 83/100 = 83% |

Accuracy as a Percentage of Total Questions

| Depth(As a percentage of accurate answers) | GPT3.5 | GPT4 |
|---|---|---|
| Prompt 1 | 4/29 ~ 13.793% | 6/34 ~ 17.647% |
| Prompt 2 | 41/45 ~ 91.111% | 44/49 ~ 89.796% |
| Totals | 45/74 ~ 60.811% | 50/83 ~ 60.241% |

Depth as a Percentage of Accurate Answers

| Interpretability(As a percentage of answers that have depth) | GPT3.5 | GPT4 |
|---|---|---|
| Prompt 1 | 0/4 = 0% | 2/6 ~ 33.333% |
| Prompt 2 | 21/41 ~ 51.22% | 27/44 ~ 61.364% |
| Totals | 21/45 ~ 46.667% | 29/50 = 58% |

Interpretability as a Percentage of Answers with Sufficient Depth

Upon first examination, I can immediately discard the observation made in the first comparison regarding a greater difference being felt when using engineered prompts. This is because GPT4 outperforms GPT3.5 in every category (and by a considerable margin) when using the non-engineered prompt. For example, GPT4-1 was 10% more accurate than GPT3.5-1 (68% compared to 58%, respectively) whereas GPT4-2 was only 8% more accurate than GPT3.5-2 (98% compared to 90%, respectively). Additionally, GPT4-2 scored more overall points in the depth category than GPT3.5-2 (44 and 41 points, respectively), but actually underperformed GPT3.5-2 in this category relative to its score in the accuracy category. This can be seen in the percentage representation where only 89.796% of GPT4-2's accurate responses earned the depth point, whereas, 91.111% of GPT3.5-2's accurate answers received the point for depth. In contrast, when using the first prompt, GPT3.5-1 underperforms GPT4-1 by a small margin, with GPT3.5-1 receiving 4 out of 29 points (13.793%) in this category and GPT4-1 receiving 6 out of 34 points (17.647%) in this category. Lastly, when looking at the interpretability section, it is evident that the discrepancy between the models in each scoring section (relative to the total amount of currently available points) is greater when using the non-engineered prompt. In this section, GPT3.5-1 scores a total of 0 out of 4 points (0%), whereas GPT4-1 scores a total of 2 out of 6 points (33.33%). In contrast, GPT3.5-2 scores a total of 21 out of 41 points (51.22%) in this section and GPT4-2 scores a total of 27/44 points (61.364%) in this section. Demonstrating that, although the nominal point difference may be greater when using the engineered prompt, the discrepancy in quality-of-response is truly felt more when using a non-engineered prompt. This finding has implications for the day-to-day user as it means that GPT4 has overall improved reasoning capabilities than GPT3.5 and is also more accessible, with the biggest discrepancy in reasoning capabilities being found at the highest level of prompt-engineering.

## 5. Discussion/Conclusion

I started this survey with the purpose of quantifying the improvement made in general reasoning capacity as well as accessibility between GPT3.5 and GPT4. Improving the capabilities of Large Language Models would be an incredible milestone for research on Natural Language processing. In turn, since natural language processing techniques (and tools) are used in bioinformatics (a crucial part of biomedicine), it would stand to reason that such an improvement could be wholly beneficial to the field of biomedicine. Additionally, with the progress that has been made in the augmentation of datasets as well as learning from live feedback (for few-shot and zero-shot scenarios), a model with high reasoning capacity could be combined with few-shot training algorithms to: diagnose patients, detect 3-dimensional patterns, generate sequences, synthesize computer programs, and run simulations on thousands of combinations. In addition to the reasoning capacity of LLMs, the effects of prompt engineering must be kept in mind as prompt engineering could further improve the performance of Large Language Models (as seen with the few-shot scenario prompts engineered in the ILF paper (Chen et al., 2024)).

In my survey, I expected to see a large difference between the overall performance of each model given both prompts (66-88 points for GPT3.5 and 209-244 points for GPT4). However, the result was extraordinarily different from my expectations; GPT3.5's total score was only 22 points less than that of GPT4 (140 for GPT3.5 and 162 for GPT4).

From my analysis of the data, I can conclude that this discrepancy between my expectations and what I observed in my data occurred for two reasons. Firstly, I did not attribute enough credit to the difference prompt-engineering would make; secondly, I overestimated the difference in performance between the two models. This finding has two implications with regard to the purpose of this paper. Firstly, the data suggests that, although GPT4 does have improved reasoning capabilities over GPT3.5, the improvement in reasoning between the two models was limited. Additionally, GPT4 was shown to perform much better with open-ended questions when given an engineered prompt. A finding that does not match the 2023 paper's (Gong et al., 2023) findings, which used multiple choice questions and concluded that GPT4's performance was relatively similar given different prompts. This difference between the findings is due to the nature of the questions as well as the scoring system, showing that GPT4 performs better and more consistently with questions that have a definitive answer. Additionally, this difference suggests that GPT4 struggles (disproportionately) more with open-ended questions that require reasoning.

Although GPT4 underperformed its overall expectations in the reasoning category, it was shown to have improved in reasoning over GPT3.5. Similarly, GPT4 has also (very subtly) improved in accessibility over GPT3.5. This is demonstrated in my results for the third comparison, where it is clearly shown that there was a greater (relative) discrepancy between the responses of GPT3.5 when given the two different prompts as compared to GPT4. This finding demonstrates that prompt-engineering, although still incredibly valuable, has less of an effect of GPT4, as it is more consistent and has a more accurate baseline than GPT3.5. This leads to the increase in its overall accuracy, while improving its accessibility to the average user.

Lastly, I want to draw a new comparison in the data to show, once again, that GPT4 is more accessible than GPT3.5 alongside being more accurate and reliable. I will be using the second set of tables for this comparison. Firstly, looking at the accuracy table, the discrepancy between GPT3.5-1 and GPT3.5-2 is 32% (58% and 90%, respectively), whereas the discrepancy between GPT4-1 and GPT4-2 is only 30%, showing that GPT4's accuracy is more consistent among different prompts. Next, the percentage of GPT3.5-1's accurate answers that were deemed to have sufficient depth was 13.793%, which is significantly lower than GPT3.5-2 with 91.111% (a difference of 77.318%). In contrast, the percentage of GPT4-1's accurate responses that were deemed to have sufficient depth was 17.647%, which is far closer to GPT4-2's 89.796% (a difference of 72.149%). Lastly, the percentage of GPT3.5-1's deep answers that were given the point for interpretability was 0% and 51.22% for GPT3.5-2 (a difference of 51.22%). On the other hand, the percentage of GPT4-1's deep answers that got the point for interpretability was 33.333%; not nearly as large of a discrepancy with GPT4-2's 61.364% (a difference of 28.031%). This data shows that GPT4 is more consistent with its responses, given the two prompts, in every category, making it more accessible to the average user. The practical effect of this finding is improvement in the accuracy of responses the average user would receive as most people don't use prompt engineering techniques and would benefit from added accessibility to LLMs.